\documentclass[runningheads]{llncs}
\usepackage{cite}
\usepackage{amsmath,amssymb,amsfonts}
\usepackage{fontawesome5}
\usepackage{algorithm}
\usepackage{algorithmic}
\usepackage{graphicx}
\usepackage{textcomp}
\usepackage{multirow} 
\usepackage{makecell} 
\usepackage{booktabs}
\usepackage{threeparttable}
\usepackage{bbm}
\usepackage{tabularx}
\usepackage{mathtools}
\usepackage{bbold}
\usepackage{caption}
\usepackage{marvosym}
\usepackage{ifsym}
\usepackage[dvipsnames]{xcolor}        

\usepackage[colorlinks,
            linkcolor=red,
            anchorcolor=red,
            citecolor=blue,
            ]{hyperref}     

\usepackage[T1]{fontenc}

\begin{document}
\title{CoCa-CXR: \textbf{Co}ntrastive \textbf{Ca}ptioners Learn Strong Temporal Structures for Chest X-Ray Vision-Language Understanding}
\titlerunning{CoCa-CXR}
%
\author{Yixiong Chen$^1$, Shawn Xu$^2$, Andrew Sellergren$^2$, Yossi Matias$^2$, Avinatan Hassidim$^2$, Shravya Shetty$^2$, Daniel Golden$^2$, Alan Yuille$^1$, Lin Yang$^2$}
\institute{Johns Hopkins University, Baltimore, USA}
\institute{$^1$Johns Hopkins University, $^2$Google Research \\\email{ychen646@jh.edu}}
\authorrunning{Yixiong Chen et al.}
%

%
\maketitle              
\begin{abstract}
Vision-language models have proven to be of great benefit for medical image analysis since they learn rich semantics from both images and reports. Prior efforts have focused on better alignment of image and text representations to enhance image understanding. However, though explicit reference to a prior image is common in Chest X-Ray (CXR) reports, aligning progression descriptions with the semantics differences in image pairs remains under-explored. In this work, we propose two components to address this issue. (1) A CXR report processing pipeline to extract temporal structure. It processes reports with a large language model (LLM) to separate the description and comparison contexts, and extracts fine-grained annotations from reports. (2) A contrastive captioner model for CXR, namely CoCa-CXR, to learn how to both describe images and their temporal progressions. CoCa-CXR incorporates a novel regional cross-attention module to identify local differences between paired CXR images. Extensive experiments show the superiority of CoCa-CXR on both progression analysis and report generation compared to previous methods. Notably, on MS-CXR-T progression classification, CoCa-CXR obtains 65.0\% average testing accuracy on five pulmonary conditions, outperforming the previous state-of-the-art (SOTA) model BioViL-T by 4.8\%. It also achieves a RadGraph F1 of 24.2\% on MIMIC-CXR, which is comparable to the Med-Gemini foundation model.
\end{abstract}
\section{Introduction}
Recent advances in vision-language (VL) pre-training have significantly enhanced the development of flexible and powerful models for chest X-ray (CXR) analysis. Training on both images and reports allows models to capture rich semantics, aligning medical concepts with image representations. However, CXR reports frequently include comparisons between multiple examinations \cite{johnson2019mimic}, a crucial aspect often overlooked by existing approaches \cite{huang2021gloria,boecking2022making,tanno2023consensus,tu2024towards}. Most prior work focuses on predicting findings from a single image without explicitly modeling temporal progression, limiting their ability to understand disease evolution and restricting their clinical applicability.

Recent literature \cite{bannur2023learning,chopretraining,bannur2024maira,wang2024hergen,yang2024unlocking} has began to utilize the temporal structure of the reports to train VL models that can make predictions based on multiple CXR images. The common practice is to fuse the representations of images as the joint representation to generate reports. While straightforward, this strategy does not explicitly guide models to learn temporal differences. The key challenges remain: (1) the lack of datasets with aligned image pairs and corresponding comparative descriptions, and (2) the need for model architectures that can effectively capture subtle regional changes over time.

\begin{figure}[!t]
\vspace{0cm}                          
\centering\centerline{\includegraphics[width=1.0\linewidth]{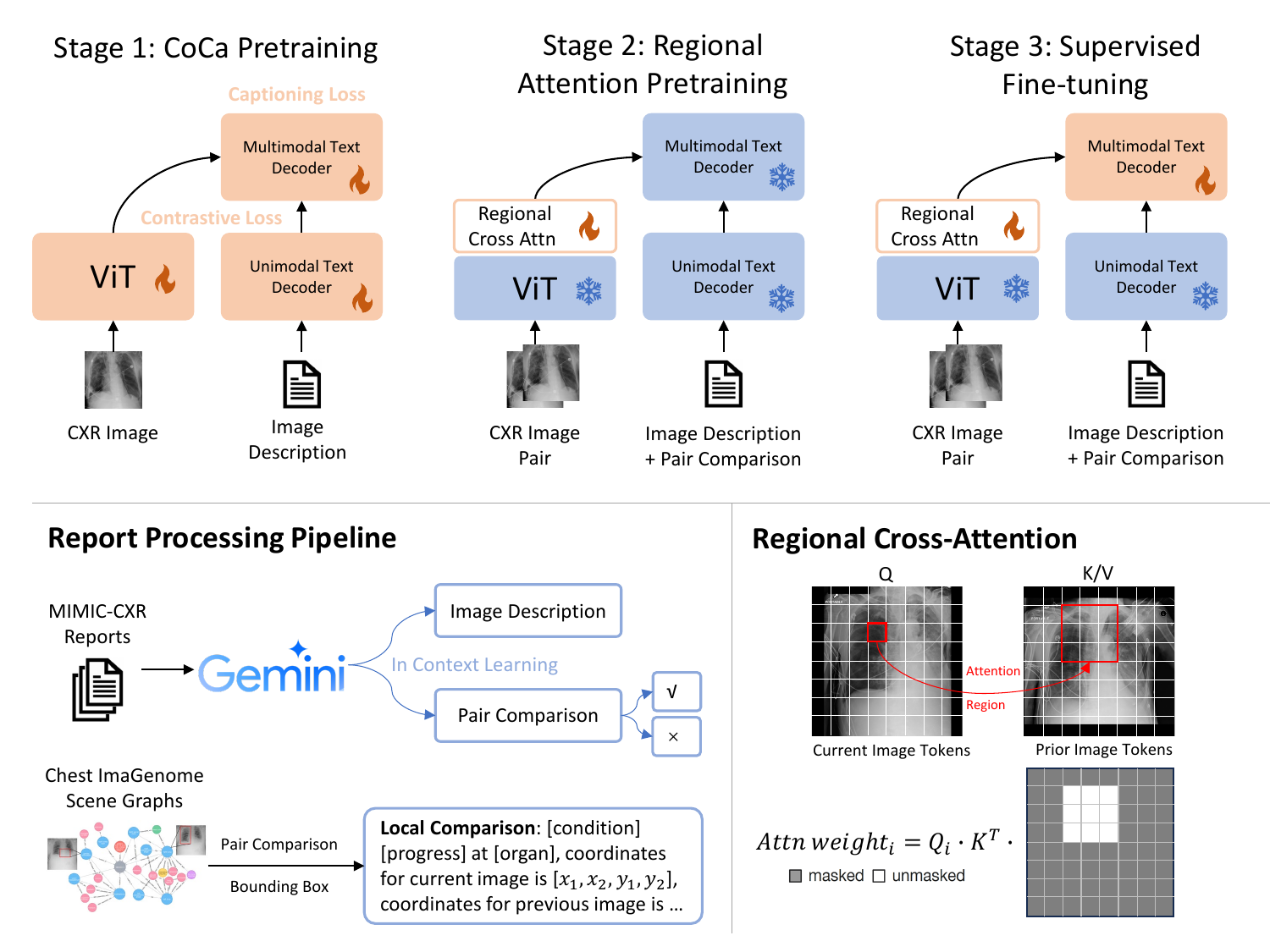}}
\caption{CoCa-CXR adds a regional cross-attention module to CoCa and is trained with three stages. We utilize an LLM, Gemini, to parse MIMIC-CXR reports and get image description and pair comparison. Chest ImaGenome scene graphs provide us with the local comparison between CXR image pairs. To leverage the locality of disease conditions, we apply cross-attention to emphasize the correlation between neighboring tokens of two images.}
\label{fig:coca_cxr}
\end{figure}

We propose a systematic framework (Fig. \ref{fig:coca_cxr}) to address the above problems. For (1), we build a CXR report processing pipeline to curate a new dataset, CXR-4, with four sub-datasets, detailed in Tab.~\ref{tab:stat}. It uses a large language model (LLM) \cite{team2023gemini} to separate reports into descriptions and comparison components, allowing the model to learn them sequentially. In addition, we also leverage the comparison and localization from scene graphs \cite{wu2021chest} which contain the description of the abnormal organs, condition progressions, and the corresponding bounding boxes, enabling the models to learn regional differences. For (2), to leverage the regional comparison in the CXR-4 dataset, we propose a regional cross-attention module, inspired by \cite{cheng2022masked}, but specifically designed for cross-attention between current and prior images. This module refines traditional cross-attention \cite{vaswani2017attention} by restricting each token in an image to attend only to its spatially corresponding tokens in the prior image. We choose Contrastive Captioner (CoCa) \cite{yucoca} as our baseline model. It has a minimalist architecture to perform VL contrastive and generative learning together, which are essential for aligned representation and downstream multitasking.

In summary, our work presents three major contributions to temporal CXR understanding. 1) We introduce the CXR-4 dataset. It provides not only the alignment between CXR images and text descriptions, but also explicit comparison. 2) We propose CoCa-CXR, which is a CoCa-based model that can generate reports from image pairs, predicting condition progressions, and localizing abnormal organs. 3) Experiments on both progression prediction and report generation tasks show the superior performance of CoCa-CXR to previous SOTA CXR temporal models.

\begin{table}[t]
\centering
\caption{Statistics of the sub-datasets used for training CoCa-CXR.}
\label{tab:stat}
\scriptsize
\renewcommand{\arraystretch}{1.3} 
\begin{tabular}{p{2.2cm}|c|p{5cm}|p{2.3cm}|c}
\toprule
\textbf{Sub-Dataset} & \textbf{Stage} & \textbf{Description} & \textbf{Data Source} & \textbf{\#Samples} \\
\hline
\hline
1. Clean image-report pair & 1\&2\&3 & Image-report pairs without any image pair comparison content. & MIMIC-CXR & 224,487 \\
\hline
2. Image pair \& filtered report & 2\&3 & Image pair with corresponding report. All reports must contain comparison info. & MIMIC-CXR & 132,320 \\
\hline
3. Image pair \& comparison info & 2\&3 & Image pair with corresponding comparison sentences from the report. & MIMIC-CXR & 259,562 \\
\hline
4. Image pair \& abnormal organs & 2\&3 & Image pair with abnormal organ's condition, coordinates, and progression. & MIMIC-CXR \& Chest ImaGenome & 758,344 \\
\bottomrule
\end{tabular}
\end{table}

\section{CXR-4 Dataset}
\label{sect:dataset}

\begin{figure}[!t]
    \centering
    \includegraphics[width=1.0\linewidth]{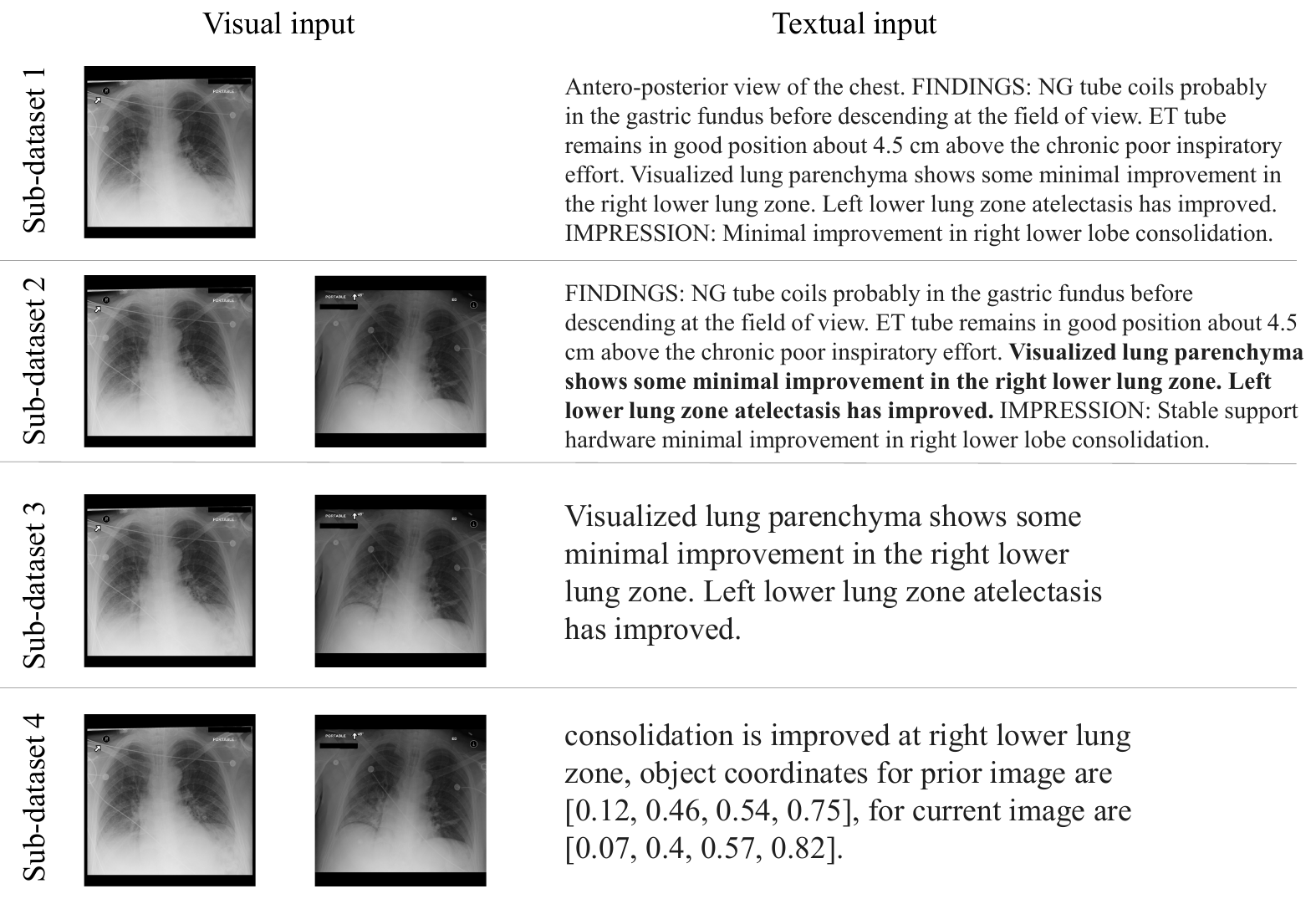}
    \caption{The illustration of the four sub-datasets of CXR-4.}
    \label{fig:enter-label}
\end{figure}

We introduce CXR-4, a new CXR dataset comprising four sub-datasets (Tab.~\ref{tab:stat}), built from MIMIC-CXR \cite{goldberger2000physiobank,johnson2019mimic,physiomimic} images, reports, and Chest ImaGenome scene graphs \cite{wu2021chest,physiochest}. The dataset follows the official MIMIC-CXR split, excluding the MS-CXR-T \cite{bannur2023learning,physiocxrt} test set. We develop a report processing pipeline (Fig.~\ref{fig:coca_cxr}) to extract structured information. Example prompts for Gemini \cite{team2023gemini} in-context learning and sample illustrations are provided in Fig.~\ref{fig:enter-label} (supplementary material). Below, we detail the sub-datasets.

\textbf{1. Clean image-report pairs.}  
To align VL representations, we pair MIMIC-CXR images with their radiological reports. We retain only AP/PA view scans and use Gemini to filter reports, keeping only view information, FINDINGS, and IMPRESSION sections while removing comparative descriptions.

\textbf{2. Image pairs \& filtered reports.}  
This subset contains samples with explicit comparison. Each sample consists of an image, its most recent prior image, and its FINDINGS and IMPRESSION sections as textual descriptions. 

\textbf{3. Image pairs \& comparative descriptions.}  
Building on (2), this subset includes sentences explicitly describing progression, extracted via Gemini. To augment the data, we reverse image pairs and modify text descriptions accordingly (e.g., “improved pneumonia” $\rightarrow$ “worsened pneumonia”).

\textbf{4. Image pair \& abnormal organs.} Using Chest ImaGenome, we extract structured comparative descriptions, providing localized abnormality progression. The format follows: "[condition] [progress] at [organ], coordinates for current image is $[x_{cur,1},x_{cur,2}$, $y_{cur,1},y_{cur,2}]$, coordinates for previous image is $[x_{prior,1},x_{prior,2}$, $y_{prior,1},y_{prior,2}]$." As additional augmentation, we reverse image pairs and adjust descriptions accordingly.

\section{CoCa-CXR}
\label{sect:method}

\subsection{Contrastive Captioners for CXR Image Pair Understanding}

CoCa \cite{yucoca} (top left in Fig.~\ref{fig:coca_cxr}) encodes images and text into latent representations using a Vision Transformer (ViT) \cite{dosovitskiy2020image} and a unimodal text encoder. A transformer text decoder then cross-attends to image features to generate captions. The model is trained with a multi-modal contrastive loss:
$$\mathcal{L}_{\text{Con}} = -\frac{1}{N} \left( \sum_{i=1}^{N} \log \frac{\exp(x_i^{\top} y_i / \tau)}{\sum_{j=1}^{N} \exp(x_i^{\top} y_j / \tau)} + \sum_{i=1}^{N} \log \frac{\exp(y_i^{\top} x_i / \tau)}{\sum_{j=1}^{N} \exp(y_i^{\top} x_j / \tau)} \right)
$$
where $x_i$ and $y_i$ are normalized image and text embeddings, $N$ is the batch size, and $\tau$ is a temperature parameter. CoCa also learns fine-grained representations through its text decoder with an autoregressive captioning objective:
$$\mathcal{L}_{\text{Cap}} = -\sum_{t=1}^{T} \log P_{\theta} (y_t | y_{<t}, x).$$
The final training objective combines both losses:
$$\mathcal{L}_{\text{CoCa}} = \mathcal{L}_{\text{Con}} + \lambda \mathcal{L}_{\text{Cap}}.$$

To adapt CoCa for modeling CXR image pairs and their temporal differences, we use its ViT encoder to extract embeddings for the current ($z^c$) and prior ($z^p$) images, where $z \in \mathbb{R}^{N \times d}$, with $N$ as the sequence length and $d$ as the feature dimension. The regional cross-attention module (detailed in Sec.~\ref{sec:crossattention}) processes $z^c$ as the main input and $z^p$ as the cross-attention input, producing an output embedding $z^o \in \mathbb{R}^{N \times d}$. We then concatenate these embeddings to form the final visual token sequence:
$$z^{concat} = concat(z^c, z^p, z^o) \in \mathbb{R}^{3N\times d}.$$
This modified architecture, CoCa-CXR, enables the model to capture both individual image semantics and temporal progression between CXR image pairs.

\subsection{Three-Stage Training of CoCa-CXR}

Training CoCa-CXR in a single stage proved challenging (see Sec.~\ref{sec:eval} ablation study), so we adopt a three-stage training strategy (Fig.~\ref{fig:coca_cxr}).

\textbf{Stage 1: CoCa pretraining.} 
We first train CoCa on the clean image-report pairs to establish a general alignment between language and visual patterns.

\textbf{Stage 2: Regional attention pretraining.}
Next, we train the regional cross-attention module using image pairs. This step is crucial because the module is randomly initialized, unlike the pretrained backbone. Prior work \cite{li2024llava,chen2023metalr} shows that prioritizing less transferable parameters improves performance.

\textbf{Stage 3: Supervised fine-tuning.} Finally, we fine-tune the regional cross-attention module and multi-modal text decoder using all four sub-datasets, refining the model’s ability to capture temporal progression and generate reports.

\subsection{Regional Cross-Attention Module}\label{sec:crossattention}

The regional cross-attention module is a Transformer block \cite{vaswani2017attention} that processes embeddings from the current ($z^c$) and prior ($z^p$) CXR images. First, a self-attention layer refines $z^c$:
\[
z^{c'} = \text{SelfAttention}(z^{c}) = \text{softmax} \left( \frac{Q_{c} K_{c}^T}{\sqrt{d}} \right) V_{c},
\]
where \( Q_c = z^c W_Q \), \( K_c = z^c W_K \), and \( V_c = z^c W_V \) are the query, key, and value projections, with learnable weights \( W_Q, W_K, W_V \).

Next, a cross-attention layer with \textbf{regional masking} (Fig.~\ref{fig:coca_cxr}, bottom right) extracts localized differences between images. The query \( Q'_i = z^{c'} W_Q' \) from the current image attends to a restricted set of key-value pairs from the prior image:
\[
\text{CrossAttention}_{i} = \text{softmax} \left( \frac{Q'_i {K'}_{\text{region}(i)}^{T}}{\sqrt{d}} \right) V'_{\text{region}(i)},
\]
where \( K'_{\text{region}(i)} = K' \odot M_{\text{region}(i)} \) and \( V'_{\text{region}(i)} = V' \odot M_{\text{region}(i)} \) are masked key-value pairs within a local window around $Q'_i$. The mask $M_{\text{region}(i)}$ selects relevant regions in the prior image.

The final output sequence \( z^o \in \mathbb{R}^{N \times d} \) is obtained by passing $\{\text{CrossAttention}_i\}_{i=1}^{N}$ through a feed-forward network, followed by skip connections and normalization.

\section{Experiments \& Results}

\subsection{Experimental Setting}

For all training stages, CXR images are padded to square, resized to $768\times768$ pixels, and normalized to [0,1] without additional augmentations. The CoCa image encoder extracts $48^2 = 2304$ visual tokens per image, which are processed by the regional cross-attention module using an $11^2$ masking window. The resulting sequence is downsampled via 2D average pooling to $16^2 = 256$ tokens before entering the multi-modal text decoder.
We train with AdamW, using a learning rate of $2\times10^{-5}$ for stages 1 and 3, and $10^{-4}$ for stage 2. The model is optimized with a batch size of $N=64$ for 20k, 10k, and 30k iterations in each stage, respectively. The sub-dataset ratio for the last two stages is set to $0.2:0.25:0.25:0.3$.

\subsection{Evalutation of CoCa-CXR}
\label{sec:eval}

\begin{table}[t]
\centering
\caption{Comparison on MS-CXR-T temporal image classification dataset (repeated for 4 random seeds). We report the macro-accuracy (\%) across the three progression classes (worsened, unchanged, improved) for each condition following BioViL-T. BioViL-T does not explicitly discuss its validation set. For a complete comparison, we report both MS-CXR-T validation and testing performance.}
\label{tab:mscxrt_comp}
\scriptsize
\renewcommand{\arraystretch}{1.5} 
\begin{tabular}{l|c|c|c|c|c|c}
\toprule
\textbf{Method} & \textbf{Consolidation} & \makecell{\textbf{Pleural}\\\textbf{effusion}} & \textbf{Pneumonia} & \textbf{Pneumothorax} & \textbf{Edema} & \textbf{Avg} \\
\hline
\hline
CNN + Transformer \cite{bannur2023learning} & 44.0 $\pm$ 2.0 & 61.3 $\pm$ 1.6 & 45.1 $\pm$ 3.5 & 31.5 $\pm$ 3.1 & 65.5 $\pm$ 1.1 & 49.5 \\
CheXRelNet \cite{karwande2022chexrelnet} & 47 & 47 & 47 & 36 & 49 & 45.2 \\
BioViL \cite{boecking2022making} & 56.0 $\pm$ 1.5 & 63.0 $\pm$ 0.9 & 60.2 $\pm$ 0.7 & 42.5 $\pm$ 2.7 & 67.5 $\pm$ 0.9 & 57.8 \\
BioViL-T \cite{bannur2023learning} & 61.1 $\pm$ 2.4 & 67.0 $\pm$ 0.8 & 61.9 $\pm$ 1.9 & 42.6 $\pm$ 1.6 & 68.5 $\pm$ 0.8 & 60.2 \\
Med-ST \cite{yang2024unlocking} & 60.6 $\pm$ 1.2 & 67.4 $\pm$ 0.3 & 58.5 $\pm$ 1.5 & 65.0 $\pm$ 0.3 & 54.2 $\pm$ 0.8 & 61.1 \\
\hline
CoCa-CXR (val.) & 70.4 $\pm$ 0.5 & 69.6 $\pm$ 1.7 & 61.4 $\pm$ 1.6 & 72.8 $\pm$ 1.1 & 71.8 $\pm$ 0.3 & 69.2 \\
CoCa-CXR (test) & 69.6 $\pm$ 2.5 & 68.1 $\pm$ 1.5 & 56.4 $\pm$ 0.8  & 59.3 $\pm$ 2.6  & 71.8 $\pm$ 0.8 & 65.0 \\
\bottomrule
\end{tabular}
\end{table}

\textbf{Results on Temporal Classification.}  
On the MS-CXR-T dataset \cite{bannur2023learning}, CoCa-CXR predicts condition progression between two images using the prompt "[condition] is ", selecting the most probable next token from \{"worsened", "unchanged", "improved"\}. As shown in Tab.~\ref{tab:mscxrt_comp}, CoCa-CXR outperforms previous SOTA models on both validation and test sets. It surpasses BioVil-T in four out of five conditions, achieving an average test accuracy of 65.0\%, which is 4.8\% higher than BioVil-T.

\begin{figure}[!t]
\vspace{0cm}                          
\centering\centerline{\includegraphics[width=1.0\linewidth]{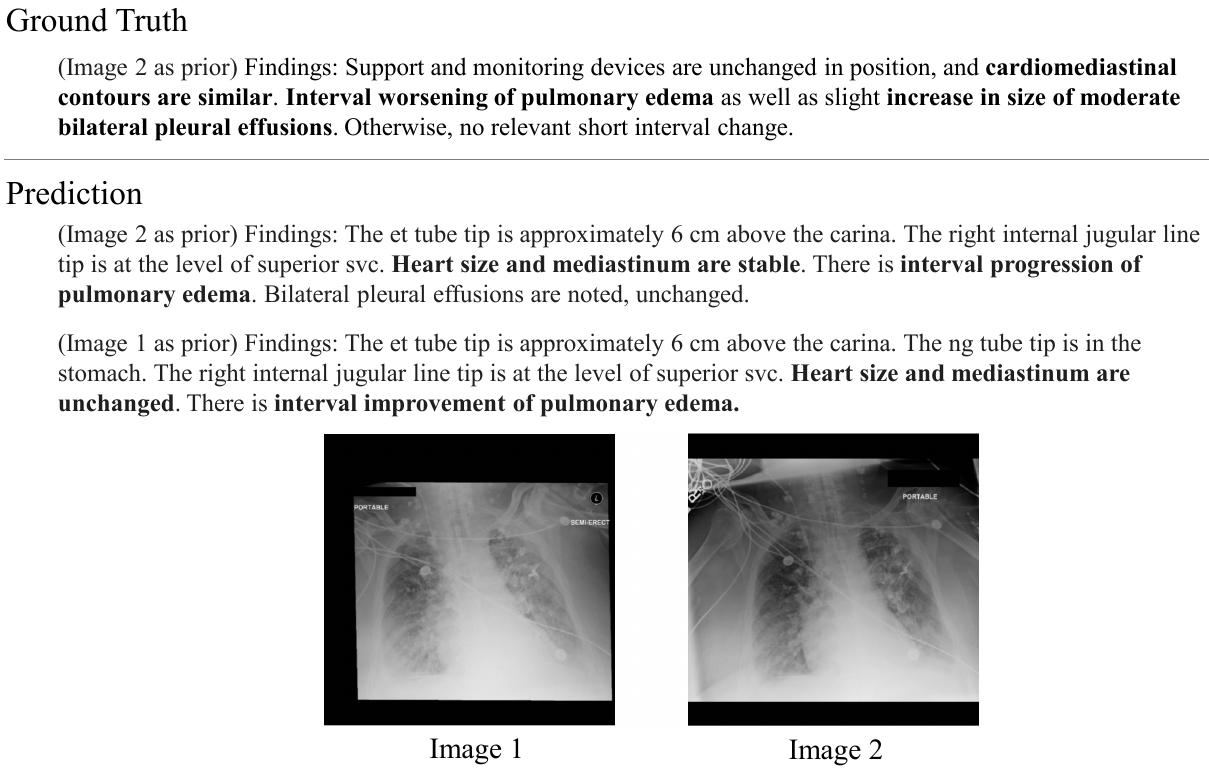}}
\caption{Report generation of CoCa-CXR on MIMIC-CXR validation set. If we swap the order of the image pair, the comparison prediction changes accordingly.}
\label{fig:pos_vs_neg_order}
\vspace{-0.0cm}
\end{figure}

\textbf{Results on Report Generation.}  
Tab.~\ref{tab:mimic-cxr} presents report generation results on the MIMIC-CXR dataset \cite{johnson2019mimic}. CoCa-CXR predicts both FINDINGS and IMPRESSION sections, a more challenging task than FINDINGS alone \cite{bannur2023learning}. When generating only descriptions, it achieves a RadGraph F1 score of 24.2\%, on par with large-scale models like Med-Gemini \cite{yang2024advancing}. Notably, PaliGemma-2 \cite{steiner2024paligemma} incorporates both images and the indication section as inputs, whereas CoCa-CXR and other models in Tab.~\ref{tab:mimic-cxr} rely solely on images; hence, we exclude its result for fair comparison.  
When also generating comparison descriptions, CoCa-CXR attains a RadGraph F1 score of 23.7\%. Although this is slightly lower than description-only generation, our hypothesis is that, describing progression adds another dimension to report generation tasks, and with the current accuracy-level (65\%) on temporal classification, one can obtain a similar level of benefit on Radgraph F1 by simply omitting comparisons in the generated reports. But still, as shown in Fig. \ref{fig:pos_vs_neg_order}, this is step forward in incorporating this progression dimension and towards real world application.

\begin{table}[!t]
\centering
\caption{CXR report generation on the MIMIC-CXR dataset
with metrics sourced from published research.}
\label{tab:mimic-cxr}
\scriptsize
\renewcommand{\arraystretch}{1.3} 

\begin{tabular}{l|l|c}
\toprule
\textbf{Method} & \textbf{Section} & \textbf{RadGraph F1 (\%)} \\
\hline
CXR-RePaiR \cite{endo2021retrieval} & Findings & 9.1 \\
$M^2$ Transformer \cite{miura2021improving} & Findings & 22.0 \\
Med-PaLM M, 12B \cite{tu2024towards} & Findings & 25.2 \\
CvT-21DistillGPT2 \cite{nicolson2023improving} & Findings $+$ Impression & 15.4 \\
Flamingo-CXR \cite{tanno2023consensus} & Findings $+$ Impression & 20.5 \\
Med-Gemini-2D \cite{yang2024advancing} & Findings $+$ Impression & 24.4 \\
\hline
CoCa-CXR (description only) & Findings $+$ Impression & 24.2 \\
CoCa-CXR (description $+$ comparison) & Findings $+$ Impression & 23.7 \\
\bottomrule
\end{tabular}
\end{table}

\begin{table}[t]
\centering
\caption{Ablation study on dataset construction, attention module, and the model training scheme. We report testing accuracy (\%) on MS-CXR-T dataset.}
\label{tab:ablation}
\scriptsize
\renewcommand{\arraystretch}{1.5} 
\begin{tabular}{c|l|c|c|c|c|c|c}
\toprule
& \textbf{Ablation} & \textbf{Con.} & \textbf{Pl. Eff.} & \textbf{Pneumon.} & \textbf{Pneumoth.} & \textbf{Edema} & \textbf{Avg} \\
\hline
 & CoCa-CXR & 69.6 & 68.1 & 56.4 & 59.3 & 71.8 & 65.0 \\
\hline
\hline
\multirow{4}{*}{\rotatebox{90}{Dataset}} & w/o Cleaning single image description & 64.8 & 70.0 & 58.5 & 56.4 & 68.4 & 63.6 \\
 & w/o Filtering comparing pairs & 65.2 & 71.2 & 54.9 & 57.9 & 70.9 & 64.0 \\
 & w/o Comparison-only description & 59.8 & 65.3 & 60.6 & 47.7 & 69.9 & 60.7 \\
 & w/o Abnormal organs \& coordinates & 54.2 & 69.1 & 59.9 & 53.2 & 65.9 & 60.5 \\
\hline
\multirow{4}{*}{\rotatebox{90}{Model}} & w/o Regional cross-attention & 58.8 & 70.5 & 58.8 & 47.2 & 69.6 & 61.0 \\
& w/o Contrastive learning & 57.4 & 68.5 & 49.8 & 45.0 & 70.7 & 58.3 \\
& w/o Stage 2 pretraining & 61.3 & 69.1 & 55.8 & 52.4 & 67.9 & 61.3 \\
& w/o Stage 1 and 2 pretraining & 58.5 & 65.5 & 58.9 & 46.2 & 62.9 & 58.4 \\
\bottomrule
\end{tabular}
\end{table}

\textbf{Ablation Study.}  
To assess the impact of CoCa-CXR’s components, we perform ablation studies on dataset construction and model training (Tab.~\ref{tab:ablation}). For temporal classification, we find that the comparison-only descriptions in sub-dataset 3 and the abnormal organ annotations with coordinates in sub-dataset 4 are crucial. The proposed regional cross-attention module improves average accuracy from 61.0\% to 65.0\%, demonstrating its effectiveness in capturing temporal differences. Additionally, contrastive loss enhances representation learning for identifying image variations. Finally, both stage 1 and stage 2 pretraining—initializing encoders and the regional cross-attention module—are essential, highlighting the importance of our three-stage training strategy.

\begin{figure}[!t]
\vspace{0cm}                          
\centering\centerline{\includegraphics[width=1.0\linewidth]{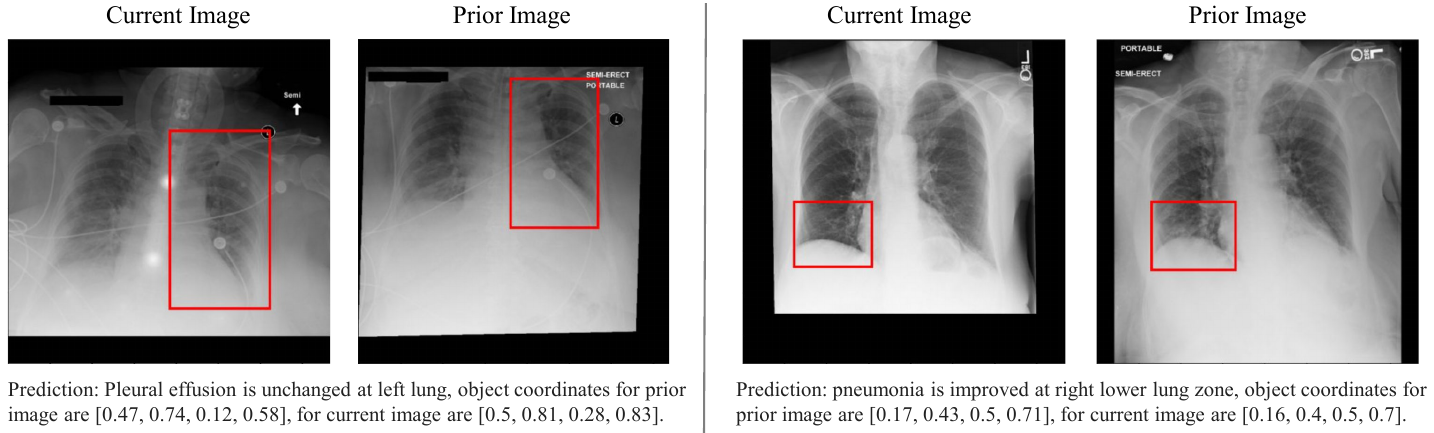}}
\caption{Visualization of the text-based condition progression detection.}
\label{fig:det}
\vspace{-0.0cm}
\end{figure}

\textbf{Visualization.} 
We visualize the learned capability of CoCa-CXR through its generated report and abnormality detection. The Fig. \ref{fig:pos_vs_neg_order} shows that the generated report can correctly describe the image content and the change from prior to current image. After swapping the order of two images, the prediction also reverse. In Fig. \ref{fig:det}, CoCa-CXR predicts the condition, progression, and the coordinates in two images demonstrating the model's capability of localizing abnormal organs. Specifically, the Intersection over Union (IoU) for Left lower lung is 0.589 on the validation set for sub-dataset 4.
A full performance breakdown on 10 pulmonary structures is provided in the supplementary material. 
These results highlight the role of vision-language alignment pretraining and regional cross-attention in capturing localized CXR patterns.

\section{Conclusion}

This work demonstrates how leveraging an LLM to curate condensed temporal information (CXR-4) enhances the training of a temporally aware model, CoCa-CXR. We introduce a regional cross-attention module to improve longitudinal CXR analysis by guiding attention across time. CoCa-CXR surpasses previous SOTA in temporal classification by incorporating explicit comparison supervision and regional attention within a three-stage training framework. It accurately predicts disease progression and generates reports with RadGraph F1 scores comparable to leading models.

\bibliographystyle{splncs04}
\bibliography{CoCaMed}

\newpage
\section{Supplementary}
\begin{table}[!htbp]
\small
\centering
\caption{Prompts for extracting comparisons.}
\setlength{\tabcolsep}{1pt}
\renewcommand{\arraystretch}{1.3} 
\begin{tabularx}{\textwidth}{lX}
\toprule
\makebox[0.20\textwidth][c]{Description} & \makebox[0.80\textwidth][c]{Prompt} \\
\hline
\hline
Extract comparison & You are a medical expert, given a report, reverse the comparison (worsen --> improved, new --> resolved, unchanged --> unchanged). If a sentence only describes the current status, remove that sentence. Following the examples below:\newline

*Example 1*\newline
*Report:*\newline
Multiple clips are again seen projecting over the left breast. Remote left-sided rib fractures are also re-demonstrated.\newline
*Reversed comparison:*\newline
Multiple clips are again seen projecting over the left breast. Remote left-sided rib fractures are also re-demonstrated.\newline
<END>\newline

*Example 2*\newline
*Report:*\newline
There is new mild pulmonary edema with small bilateral pleural effusions. Lung volumes have decreased.\newline
*Reversed comparison:*\newline
Mild pulmonary edema has resolved. Lung volumes have increased.\newline
<END>\newline

*Example 3*\newline
*Report:*\newline
Unexplained mild rightward deviation of the trachea without tracheal narrowing at the level of the thoracic inlet, not markedly changed since \_\_\_. No change in the probable right apical bronchiectasis.\newline
*Reversed comparison:*\newline
Unexplained mild rightward deviation of the trachea without tracheal narrowing at the level of the thoracic inlet, not markedly changed since \_\_\_. No change in the probable right apical bronchiectasis.\newline
<END>\newline

*Example 4*\newline
*Report:* [Original Report]
\\

\bottomrule
\end{tabularx}
\end{table}

\begin{table}[!htbp]
\small
\centering
\caption{Prompts for cleaning reports.}
\setlength{\tabcolsep}{1pt}
\renewcommand{\arraystretch}{1.3} 
\begin{tabularx}{\textwidth}{lX}
\toprule
\makebox[0.25\textwidth][c]{Description} & \makebox[0.75\textwidth][c]{Prompt} \\
\hline
\hline
Clean reports & You are a medical expert, given a Chest X-ray report, please rewrite the FINDINGS section and IMPRESSION section to only contain information that can be noted in the current chest X-ray image (i.e. no comparison with prior chest X-ray image, no referencing to prior or CT images). For findings you can infer current status from the comparison, rewrite them, for example, rewrite "worsening pleural effusion" to "pleural effusion is present". For things you can not infer the current status, for example, "mediastinal contours are unchanged", drop that finding.\newline
Following the examples below:\newline

**Example 1**\newline
*Report:*\newline
INDICATION: Female with shortness of breath.
FINDINGS: Single portable view of the chest is compared to previous exam from \_\_\_. Enteric tube is seen with tip off the inferior field of view. Left PICC is seen; however, tip is not clearly delineated. Persistent bibasilar effusions and a right pigtail catheter projecting over the lower chest. There is possible right apical pneumothorax. Superiorly, the lungs are clear of consolidation. Cardiac silhouette is within normal limits. Osseous and soft tissue structures are unremarkable.\newline
IMPRESSION: No significant interval change with bilateral pleural effusions with right pigtail catheter in the lower chest.\newline
*Output:*\newline
INDICATION: Female with shortness of breath.
FINDINGS: Enteric tube is seen with tip off the inferior field of view. Left PICC is seen; however, tip is not clearly delineated. Bibasilar effusions and a right pigtail catheter projecting over the lower chest. There is possible right apical pneumothorax. Superiorly, the lungs are clear of consolidation. Cardiac silhouette is within normal limits. Osseous and soft tissue structures are unremarkable.
IMPRESSION: Bilateral pleural effusions with right pigtail catheter in the lower chest.\newline
<END>\newline

*Example 2*\newline
*Report:* [Original Report] \\

\bottomrule
\end{tabularx}
\end{table}

\newpage

\begin{table}[!t]
\centering
\caption{Detection performance (IoU) for the current and prior image for different organs. For small anatomical structures like costophrenic angle, the detection may not be as accurate as large organs due to their sizes and the output precision (two decimal places) of the language model.}
\renewcommand{\arraystretch}{1.0}
\begin{tabular}{l|c|c}
\toprule
    Organ  & Current image & Prior image  \\
\hline
\hline
Left apical zone & 0.591 & 0.506  \\
Left costophrenic angle & 0.182 & 0.153 \\
Left hilar structures & 0.553 & 0.472 \\
Left lower lung zone & 0.589 & 0.598 \\
Left lung & 0.726 & 0.702 \\
Right apical zone & 0.576 & 0.540  \\
Right costophrenic angle & 0.224 & 0.177 \\
Right hilar structures & 0.438 & 0.666 \\
Right lower lung zone & 0.573 & 0.583 \\
Right lung & 0.808 & 0.682 \\
Mean & 0.344 & 0.307\\
\bottomrule
\end{tabular}
\end{table}
\end{document}